\documentclass[runningheads]{llncs}
\usepackage{graphicx}
\usepackage{amsmath,amsfonts,bm}
\usepackage[caption=false]{subfig}
\usepackage{wrapfig}
\usepackage{ stmaryrd }
\usepackage[export]{adjustbox}
\usepackage{booktabs}
\usepackage{diagbox}
\usepackage{capt-of}
\usepackage{tablefootnote}
\usepackage{hyperref}
\begin{document}
\title{Unsupervised Discovery of 3D Hierarchical Structure with Generative Diffusion Features}
\titlerunning{Unsupervised Discovery of 3D Hierarchical Structure}
%
\author{Nurislam Tursynbek\and
Marc Niethammer}
%
\authorrunning{N. Tursynbek et al.}
%
\institute{University of North Carolina, Chapel Hill, NC, USA}
%
\maketitle              
\begin{abstract}
Inspired by findings that generative diffusion models learn semantically meaningful representations, we use them to discover the intrinsic hierarchical structure in biomedical 3D images using \emph{unsupervised} segmentation. We show that features of diffusion models from different stages of a U-Net-based ladder-like architecture capture different hierarchy levels in 3D biomedical images. We design three losses to train a predictive unsupervised segmentation network that encourages the decomposition of 3D volumes into meaningful nested subvolumes that represent a hierarchy. First, we pretrain 3D diffusion models and use the consistency of their features across subvolumes. Second, we use the visual consistency between subvolumes. Third, we use the invariance to photometric augmentations as a regularizer. Our models perform better than prior unsupervised structure discovery approaches on challenging biologically-inspired synthetic datasets and on a real-world brain tumor MRI dataset. Code is available at \href{https://github.com/uncbiag/diffusion-3D-discovery}{github.com/uncbiag/diffusion-3D-discovery}.

\end{abstract}

\section{Introduction}

Deep neural networks (DNNs) have been successfully applied to various supervised 3D biomedical image analysis tasks, such as classification \cite{gao2017classification}, segmentation \cite{cciccek20163d}, and registration \cite{yang2017quicksilver}. Acquiring volumetric annotations manually to supervise deep learning models is costly and labor intensive. For example, the supervised training of 3D DNNs for segmentation requires the manual labeling of every voxel of the structures of interest for the entire training set. 
Additionally, the diversity of existing biomedical 3D volumetric image types (e.g. MRI, CT, electron tomography) and different tasks associated with them precludes image annotations for all existing problems in practice. Furthermore, experts may focus on annotating objects they are already aware of, thereby restricting the possibility of new structural discoveries in large datasets using deep learning. We hypothesize that the nested hierarchical structure intrinsic to many 3D biomedical images~\cite{hsu2021capturing} might be useful for unsupervised segmentation. As a step in this direction, our goal in this work is to develop a computational approach for unsupervised structure discovery. 

\begin{figure}[t]
\centering
\begin{minipage}{0.8\linewidth}
\begin{picture}(200,142)
\put(0,0){\includegraphics[width=\linewidth]{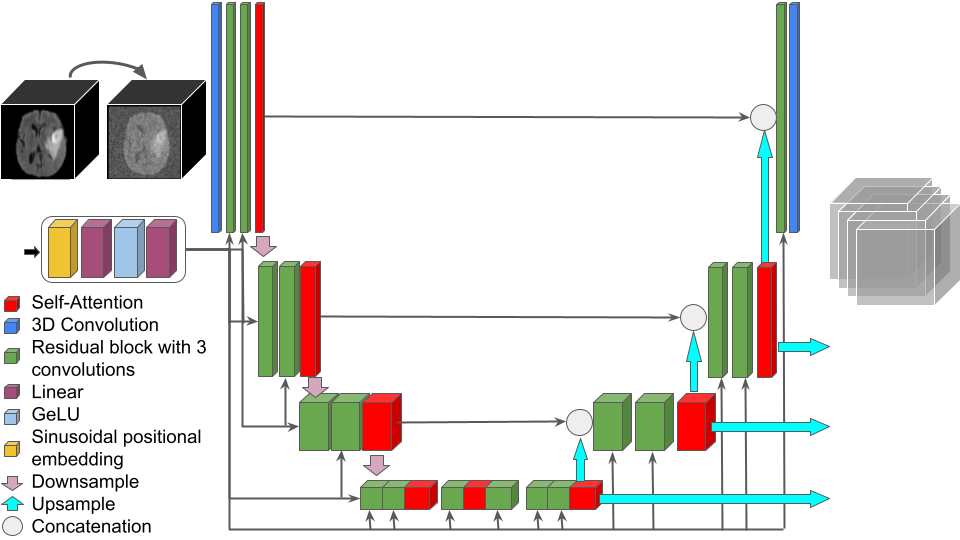}}
\put(2,81){$t$}
\put(14,142){$q(\mathbf{x}_t|\mathbf{x}_0)$}
\put(247,80){$\bm{\phi}(\mathbf{x}_0)$}
\put(244,54){Stage 3 features}
\put(244,30){Stage 2 features}
\put(244,9){Stage 1 features}
\end{picture}
\end{minipage}
\caption{\textbf{Feature extractor.} Given a clean 3D image $\mathbf{x}_0$, we add Gaussian noise corresponding to diffusion timestep $t$ to the image following the distribution $q(x_t|x_0)$ using Eq.~\eqref{eq2}. The noisy image $\mathbf{x}_t$ is passed to our pretrained 3D diffusion model. We upsample intermediate activations to the original image size and use them as feature extractor $\bm{\phi}$ for each voxel. Features from different stages of a U-Net-based ladder-like architecture for a diffusion model capture different hierarchy levels.}\vspace{-1em}
\label{fe}
\end{figure}
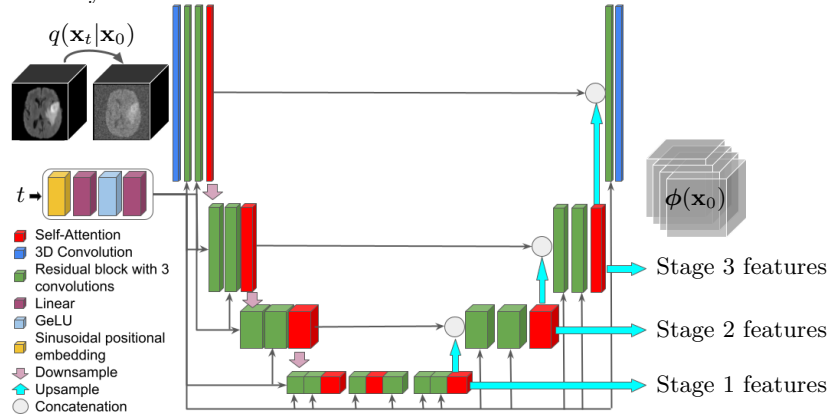

Recently, unsupervised part discovery in 2D natural images has gained significant attention~\cite{collins2018deep,hung2019scops,choudhury2021unsupervised}. These methods are based on the finding that intermediate activations of deep ImageNet-pre-trained classification models capture semantically meaningful conceptual regions~\cite{collins2018deep}. These regions are robust to pose and viewpoint variations and help high-level image understanding by providing local object representations, leading to more explainable recognition~\cite{hung2019scops}. However, a naive application of part discovery methods to 3D volumetric segmentation is not feasible, due to the lack of good feature extractors for 3D biomedical images~\cite{chen2019med3d} and ImageNet-pretrained networks operate only on 2D images.

We hypothesize that deep generative models are good feature extractors for unsupervised structure discovery for the following reasons. First, these models do not require expert labels as they are trained in a self-supervised way. Second, the ability to generate high-quality images suggests that these models capture semantically meaningful information. Third, generative representation learning has been successfully applied to global and dense prediction tasks in 2D images~\cite{donahue2019large} and has shown improvements in label efficiency and generalization~\cite{li2021semantic}. 

Besides creating stunning image generation results, diffusion-based generative models~\cite{ho2020denoising} are applied to other downstream tasks. Several works use pretrained diffusion models for 2D label-efficient semantic segmentation of natural images~\cite{baranchuk2021label,asiedu2022decoder}. In 2D medical imaging, diffusion models are used for self-supervised vessel segmentation~\cite{kim2022diffusion}, anomaly detection \cite{wolleb2022diffusion,wyatt2022anoddpm,sanchez2022healthy,pinaya2022fast}, denoising~\cite{hu2022unsupervised}, and improving supervised segmentation models~\cite{wolleb2022diffusion1,wu2022medsegdiff}. In 3D medical imaging, diffusion models are used for CT and MR image synthesis~\cite{dorjsembe2022three,wu2022medsegdiff}. Inspired by the success of unsupervised part discovery methods in 2D images and the effective abilities of diffusion models for many downstream tasks we hypothesize that feature representations of generative diffusion models discover intrinsic hierarchical structures in 3D biomedical images. Our work explores this hypothesis.

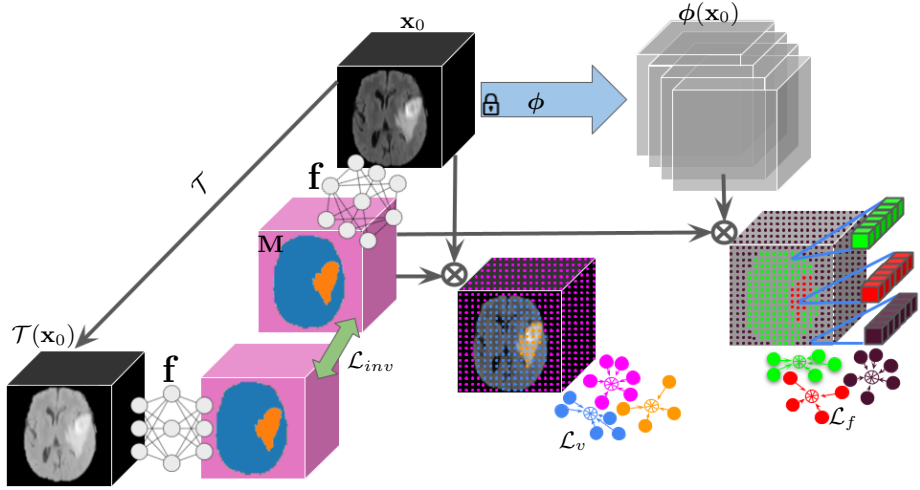
\begin{figure}[t]
\centering
\begin{minipage}{\linewidth}
\begin{picture}(200,145)
\put(0,0){\includegraphics[width=\linewidth, trim={0 1.65cm 0 0},clip]{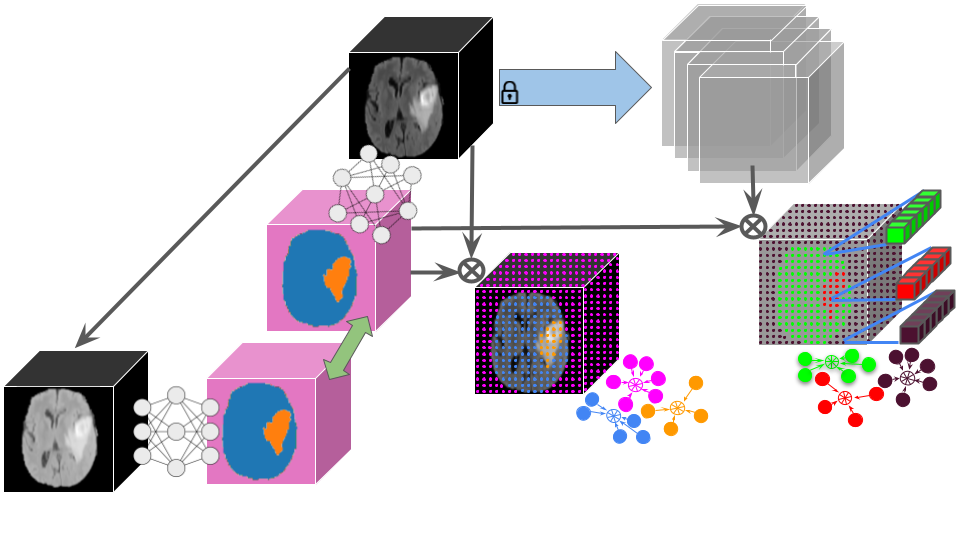}}
\put(198,143){$\bm{\phi}$}
\put(150,174){$\mathbf{x}_0$}
\put(255,177){$\bm{\phi}(\mathbf{x}_0)$}
\put(96, 89){$\mathbf{M}$}
\put(210, 15){$\mathcal{L}_v$}
\put(312, 24){$\mathcal{L}_f$}
\put(130, 45){$\mathcal{L}_{inv}$}
\put(115, 113){\Large{$\mathbf{f}$}}
\put(60, 40){\Large{$\mathbf{f}$}}
\put(70, 110){\rotatebox{45}{$\mathcal{T}$}}
\put(3, 55){$\mathcal{T}(\mathbf{x}_0)$}
\end{picture}
\end{minipage}
\caption{\textbf{Predictive unsupervised structure discovery.} Our unsupervised segmentation network $\mathbf{f}: \mathbf{x}_0\rightarrow \mathbf{M}$ is trained with three losses. Feature consistency loss $\mathcal{L}_f$ encourages features $\bm{\phi}(\mathbf{x}_0)$, extracted using diffusion models (see Fig.~\ref{fe}), of voxels belonging to the same parts to be similar to each other. Visual consistency loss $\mathcal{L}_v$ encourages models to learn parts that align with image boundaries. Photometric invariance loss $\mathcal{L}_{inv}$ encourages invariance in models to photometric transformation $\mathcal{T}$.}
\label{proc}
\end{figure}

\textbf{Our contributions are:}
\begin{itemize}
    \item[1)] We pretrain 3D diffusion models, use them as feature extractors (Fig.~\ref{fe}), and design losses (Fig.~\ref{proc}) for unsupervised 3D structure discovery. 
    \item[2)] We show that features from different stages of ladder-like U-Net-based diffusion models capture different hierarchy levels in 3D biomedical volumes.
    \item[3)] Our approach outperforms previous 3D unsupervised discovery methods on challenging synthetic datasets and on a real-world brain tumor segmentation (BraTS'19) dataset.
\end{itemize}

\section{Background on Diffusion Models}
Diffusion models \cite{ho2020denoising} consist of two parts: a forward pass and a reverse pass. The forward pass is a $T$-step process of adding a small Gaussian noise, gradually destroying image information and transforming a clean image $\mathbf{x}_0$ into pure Gaussian noise $\mathbf{x}_T$. Each step $t\in\llbracket 1, T\rrbracket$ is:
\begin{equation}
    q(\mathbf{x}_t|\mathbf{x}_{t-1}):=\mathcal{N}(\mathbf{x}_t; \sqrt{1-\beta_t}\mathbf{x}_{t-1}, \sqrt{\beta_t}\mathbf{I})\,,
\end{equation}
where $\{\beta_t\}_{t=1}^T$ is a variance schedule. With $\overline{\alpha}_t = \prod_{i=1}^t(1-\beta_i)$, the noisy image $\mathbf{x}_t$ at a timestep $t$, following  $q(\mathbf{x}_t|\mathbf{x}_{0})=\prod_{i=1}^tq(\mathbf{x}_i|\mathbf{x}_{i-1})$, can be written as:
\begin{equation}
    \mathbf{x}_t = \sqrt{\overline{\alpha}_t}\mathbf{x}_0 + \sqrt{1-\overline{\alpha}_t}\bm{\epsilon}\,, \qquad \bm{\epsilon}\sim\mathcal{N}(\mathbf{0},\mathbf{I})\,.
    \label{eq2}
\end{equation}

\noindent The reverse pass is a corresponding $T$-step denoising process using a neural network (usually, U-Net \cite{ronneberger2015u}) with parameters $\theta$. For small noises, the reverse pass is also Gaussian:

\begin{equation}
    p_\theta(\mathbf{x}_{t-1}|\mathbf{x}_t):=\mathcal{N}(\mathbf{x}_{t-1};\bm{\mu}_\theta(\mathbf{x}_t, t), \bm{\Sigma}_\theta(\mathbf{x}_t, t))\,.
\end{equation}

Practically, instead of $\bm{\mu}_\theta(x_t, t)$ and $\bm{\Sigma}_\theta(\mathbf{x}_t, t)$, models are designed to predict either the noise $\bm{\epsilon}_t$ at timestep $t$, or a less noisier version of image $\mathbf{x}_{t-1}$ directly. 

\section{Method}

We formulate the 3D structure discovery task in biomedical images as an unsupervised segmentation into $K$ parts. Given a one-channel 3D image $\mathbf{x}_0\in\mathbb{R}^{1\times H\times W\times D}$, our segmentation model $\mathbf{f}$ predicts a mask $\mathbf{M}\in[0,1]^{K\times H\times W\times D}$. For all voxels $u\in\llbracket 0,H-1\rrbracket\times\llbracket 0,W-1\rrbracket\times\llbracket 0,D-1\rrbracket$, we have $\sum_{k=1}^K\mathbf{M}_{ku}=1$. We use three losses for unsupervised training (see Fig.~\ref{proc}):
\begin{equation}
\mathcal{L}=\lambda_v\mathcal{L}_v+\lambda_f\mathcal{L}_f+\lambda_{inv}\mathcal{L}_{inv}\,.
\label{loss}
\end{equation}

For an arbitrary representation  $\bm{h}(\mathbf{x}_0)$ of an image $\mathbf{x}_0$ with voxels $u$, the consistency of this representation $C(\bm{h}(\mathbf{x}_0))$ across $K$ predicted parts in the form of segmentation $\mathbf{M}$ is defined as:
\begin{equation}
    C(\bm{h}(\mathbf{x}_0))=\frac{1}{N}\sum_{k=1}^K\sum_u\mathbf{M}_{ku}\|\mathbf{z}_k-[\bm{h}(\mathbf{x}_0)]_{u}\|_2^2,\quad\text{where}\quad \mathbf{z}_k = \frac{\sum_u \mathbf{M}_{ku}[\bm{h}(\mathbf{x}_0)]_u}{\sum_{u}\mathbf{M}_{ku}}\,,
    \label{consistency}
\end{equation}
where $N$ is the number of voxels. This is a form of volume-normalized K-means loss with $z_k$ describing the mean feature value of partition $k$.

\textbf{Feature Consistency.} We pretrain generative 3D diffusion models and use them as feature extractors \cite{baranchuk2021label}. Noise is added to a clean image $\mathbf{x}_0$ based on Eq.~\eqref{eq2} and the noisy image $\mathbf{x}_t\in\mathbb{R}^{1\times H\times W\times D}$ is passed to the 3D diffusion model. Intermediate activations (either from different stages of ladder-like U-Nets or their concatenation, see Fig.~\ref{fe}) upsampled to the original image size serve as a $p-$dimensional feature extractor $\bm{\phi}(\mathbf{x}_0)\in\mathbb{R}^{p\times H\times W\times D}$. The feature consistency loss encourages voxels corresponding to the same parts to have similar features:
\begin{equation}
    \mathcal{L}_f = C(\bm{\phi}(\mathbf{x}_0))\,. \label{lossf}
\end{equation}

\textbf{Visual Consistency.} The extracted features are upsampled from low spatial resolutions and therefore do not accurately align with image boundaries. To alleviate this problem, we use a voxel visual consistency loss:
\begin{equation}
    \mathcal{L}_v = C(\bm{I}(\mathbf{x}_0)) = C(\mathbf{x}_0)\,
    \label{lossv}
\end{equation}
where $\bm{I}(\mathbf{x}_0)$ is the identity feature extractor, i.e. $\bm{I}(\mathbf{x}_0)=\mathbf{x}_0$.

\textbf{Photometric Invariance.} As biomedical images often show acquisition differences (e.g., based on MR or CT scanner), they can be heterogeneous in their voxel intensities~\cite{nalepa2019data}. Therefore, robustness of models to voxel-level photometric perturbations might be helpful for \emph{unsupervised} discovery. We use the Dice loss~\cite{milletari2016v} to encourage invariance to such a photometric transformation $\mathcal{T}$:

\begin{equation}
    \mathcal{L}_{inv} = 1 - \frac{1}{K}\sum_{k=1}^K\frac{2\sum_u [\mathbf{f}(\mathbf{x}_0)]_{ku}[\mathbf{f}(\mathcal{T}(\mathbf{x}_0))]_{ku}}{\sum_u [\mathbf{f}(\mathbf{x}_0)]_{ku}^2+\sum_u [\mathbf{f}(\mathcal{T}(\mathbf{x}_0))]_{ku}^2}\,.
\end{equation}
We assume our images are min-max normalized ($\mathbf{x}_0\in[0,1]$). We then use gamma-correction of the form $\mathcal{T}(\mathbf{x}_0)=\mathbf{x}_0^\gamma$ as a photometric transformation. We draw $\gamma$ from the uniform distribution:  $\gamma\sim U[\gamma_{min}, \gamma_{max}]$.

\begin{table*}
\centering
\begin{tabular}{l@{\hskip 0.05in}c@{\hskip 0.05in}c@{\hskip 0.05in}c@{\hskip 0.1in}c@{\hskip 0.1in}c@{\hskip 0.05in}c@{\hskip 0.05in}c}
\toprule
 & \multicolumn{3}{c}{Regular} &  & \multicolumn{3}{c}{Irregular} \\
\cmidrule{2-4}\cmidrule{6-8}
& $Level\;1$ & $Level\;2$ & $Level\;3$ &  & $Level\;1$ & $Level\;2$ & $Level\;3$\\
\midrule
{\c{C}}i{\c{c}}ek et al \cite{cciccek20163d} & 0.968 & 0.829 & 0.668 & Semi-supervised & 0.970 & 0.825 & 0.601\\
Zhao et al \cite{zhao2019data} & 0.989 & 0.655 & 0.357 & Semi-supervised & 0.978 & 0.641 & 0.333 \\
\midrule
Nalepa et al \cite{nalepa2020unsupervised} & 0.530 & 0.276 & 0.112 & Unsupervised & 0.527 & 0.280 & 0.144 \\
Ji et al \cite{ji2019invariant} &  0.589 & 0.291 & 0.150 & Unsupervised & 0.527 & 0.280 & 0.144\\
Moriya et al \cite{moriya2018unsupervised} & 0.628 & 0.311 & 0.141 & Unsupervised & 0.525 & 0.232 & 0.094\\
Hsu et al \cite{hsu2021capturing} & 0.952 & 0.541 & 0.216 & Unsupervised & 0.953 & 0.488 & 0.199\\ 
 \textbf{Ours} & \textbf{0.986} & \textbf{0.577} & \textbf{0.397} & Unsupervised & \textbf{0.967} & \textbf{0.565} & \textbf{0.382} \\
\midrule
$k$-means & 0.808 & 0.326 & 0.149 & Non-DL & 0.771 & 0.299 & 0.118 \\
BM4D+$k$-means & 0.949 & 0.529 & 0.335 & Non-DL & 0.950 & 0.533 & 0.324 \\
\bottomrule
\end{tabular}
\caption{Dice scores on the biologically inspired synthetic datasets. Our method outperforms all previous work on unsupervised 3D segmentation for all levels of hierarchy.}
\label{regular}
\vspace{-2em}
\end{table*}

\section{Experiments}

\subsection{Datasets}

To compare with state-of-the-art unsupervised 3D segmentation methods we follow~\cite{hsu2021capturing} and evaluate our method on challenging biologically inspired 3D synthetic datasets and a real-world brain tumor segmentation (BraTS'19) dataset. 

The synthetic dataset of~\cite{hsu2021capturing}, consists of 120 volumes (80-20-20 split) of size $50\times50\times50$. Inspired by cryo-electron tomography images, it contains a three-level structure, representing a biological cell, vesicles and mitochondria, as well as protein aggregates. The intensities and locations of the objects are randomized without destroying the hierarchy. The regular variant of the dataset contains cubical and spherical objects, while the irregular variant contains more complex shapes. Pink noise of magnitude $m=0.25$ which is commonly seen in biological data~\cite{sejdic2013necessity} 
 is applied to the volume. Fig.~\ref{synthdatasets} shows sample slices of both variants. 

\begin{table}[ht]
\begin{minipage}[b]{0.45\linewidth}
\centering
\begin{tabular}{l@{\hskip 0.05in}c@{\hskip 0.05in}c@{\hskip 0.05in}c}
    \toprule
    & \multicolumn{3}{c}{Predictions}\\
    \cmidrule{2-4}
    & Level 1 & Level 2 & Level 3\\\midrule
    Stage 1 features & \textbf{0.986} & 0.366  & 0.273\\\midrule
    Stage 2 features& 0.923 & \textbf{0.577} & 0.327\\\midrule
    Stage 3 features & 0.878 & 0.489 & \textbf{0.397} \\
    \bottomrule
\end{tabular}
\caption{Dice scores when using features from different stages of ladder-like U-Net-based diffusion models (see Fig.~\ref{fe}) for unsupervised segmentation of the different hierarchy levels of the synthetic dataset. Features at lower resolutions (Stage 1) are more suitable for discovering larger objects (Level 1). Intermediate features (Stage 2) are more suitable for intermediate discoveries (Level 2). Features at higher resolutions (Stage 3) are more suitable for more detailed discoveries (Level 3).}
    \label{levels}
\end{minipage}\hfill
\begin{minipage}[b]{0.5\linewidth}
\begin{picture}(2000, 200)
\put(8,0){\includegraphics[trim=0cm 0cm 16cm 0cm, clip=true,width=0.96\linewidth]{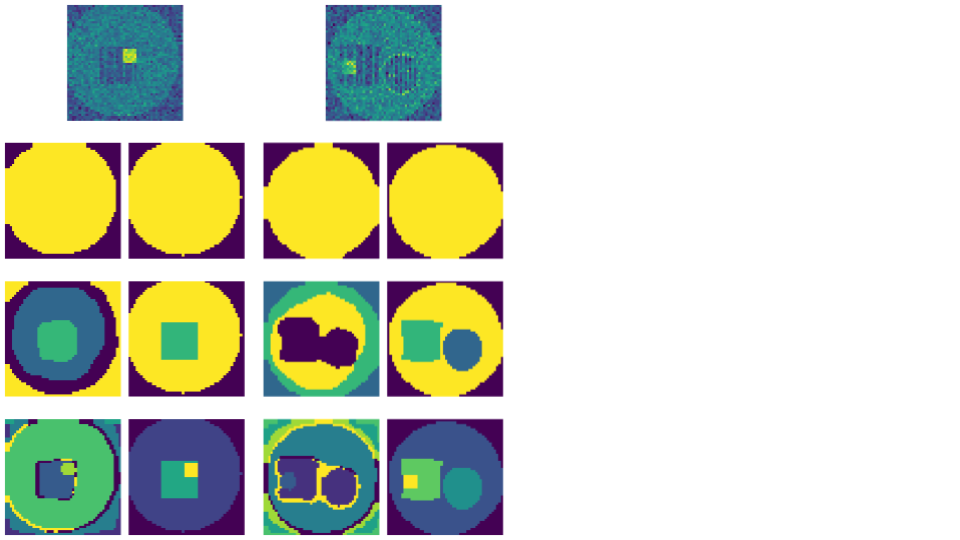}}
\put(0,4){\rotatebox{90}{Level 3}}
\put(0,50){\rotatebox{90}{Level 2}}
\put(0,98){\rotatebox{90}{Level 1}}
\put(20,180){Regular Slice}
\put(110,180){Iregular Slice}
\put(12,133){Pred.}
\put(67,133){GT}
\put(96,133){Pred.}
\put(153,133){GT}
\end{picture}
\captionof{figure}{Examples of unsupervised 3D structure discovery with our method on the biologically-inspired synthetic datasets. GT indicates ground truth.}
\label{synthdatasets}
\end{minipage}
\vspace{-2em}
\end{table}

The BraTS'19 dataset~\cite{menze2014multimodal,bakas2017advancing,bakas2018identifying} is an established benchmark for 3D tumor segmentation of brain MRIs. Volumes are co-registered to the same template, interpolated to $(1 mm)^3$ resolution and brain-extracted. Following~\cite{hsu2021capturing}, images are cropped to volumes of size $200 \times 200 \times 155$. As in~\cite{hsu2021capturing}, FLAIR images and corresponding whole tumor (WT) annotations are used for unsupervised segmentation evaluation with the same split of $259$ high grade glioma training examples into $180$ train, $39$ validation, and $40$ test samples. The official BraTS'19 validation and test sets are not used as their segmentation masks are not available.

\subsection{Implementation Details}

All diffusion models use the same architecture shown in Fig.~\ref{fe}. We pretrain them for $50k$ epochs with batch size $4$, using an $L1$ loss between the denoised and the original images. We use the Adam optimizer, a cosine noise schedule, learning rate $10^{-4}$ and $T=250$ steps. The first layer has $64$ channels and this number is doubled for the proceeding downsampling layers. Due to memory constraints for BraTS'19, we trained diffusion models at $128\times128\times128$ resolution. However, the extracted features are upsampled to the original $200\times200\times155$ resolution. 

Our segmentation networks ($\mathbf{f}$ in Fig.~\ref{proc}) use a 3D U-Net architecture~\cite{ronneberger2015u,cciccek20163d}. We trained them for $100$ epochs using the Adam optimizer, a learning rate of $3*10^{-4}$ and the losses in Eq.~\eqref{loss}. We selected the epoch that gave the best average probability of the segmentation mask for all inputs \cite{park2022encoder} as our final model. Noisy images at timestep $t=25$ are used as input to the diffusion models. Due to the memory limits, for BraTS'19, we used Stage 2 features, as they have the least number of channels. We set $\lambda_f=\lambda_v=\lambda_{inv}=1$ and $\gamma\sim U[0.9,1.1]$ for all cases. For all experiments we used Pytorch and 4 NVIDIA A6000 GPUs (48Gb).

\subsection{Results}
We compare our method with state-of-the-art unsupervised 3D structure discovery approaches including clustering using 3D feature learning~\cite{moriya2018unsupervised}, a 3D convolutional autoencoder~\cite{nalepa2020unsupervised}, and self-supervised hyperbolic representations~\cite{hsu2021capturing}.

For the synthetic datasets, we used $K=2$ (background and cell) for Level 1, $K=4$ (background, cell, vesicle, mitochondria) for Level 2, and $K=8$ (background, cell, vesicle, mitochondria, and 4 small protein aggregates) for Level 3 predictions. The evaluation metric is the average Dice score on the annotated test labels. As the label order may differ we use the Hungarian algorithm to match the predicted masks with the ground truth segmentations. Tab.~\ref{regular} shows the results for the regular and irregular variants of the cryo-ET-inspired synthetic dataset. Our models outperform all previous unsupervised work at all hierarchy levels. For some levels, our models even outperform semi-supervised methods ({\c{C}}i{\c{c}}ek et al.~\cite{cciccek20163d} used $2\%$ of annotated data, Zhao et al.~\cite{zhao2019data} used one annotated volume). We found that simple unsupervised denoising (BM4D \cite{maggioni2012nonlocal}) followed by $k$-means clustering provides a good baseline, although vanilla k-means clustering on voxel intensities does not perform well due to noise. Results in Fig.~\ref{synthdatasets} demonstrate that our proposed unsupervised method indeed discovers the hierarchical structure of different levels. We also show in Tab.~\ref{levels} that features from early decoder stages of the U-Net-based diffusion models better discover larger objects in the hierarchy, features at intermediate stages better capture intermediate objects, and features at later stages better find smaller objects.

\begin{table}[t]
    \centering
    \begin{tabular}{l@{\hskip 0.1in}@{\hskip 0.1in}c@{\hskip 0.1in}@{\hskip 0.1in}c@{\hskip 0.1in}@{\hskip 0.1in}c}
\toprule 
& $Dice\; WT\uparrow$ & $HD95\; WT\downarrow$  & \\
\midrule
1st place solution \cite{jiang2019two} & 0.888 & 4.618 &  Supervised\\
\midrule
\tablefootnote{Dice and HD95 numbers for these models are taken from \cite{hsu2021capturing}\label{note1}.} Nalepa et al \cite{nalepa2020unsupervised} & 0.211 & 170.434 & Unsupervised\\
\textsuperscript{\ref{note1}} Ji et al \cite{ji2019invariant} &  0.425 & 114.400 &  Unsupervised\\
\textsuperscript{\ref{note1}} Moriya et al \cite{moriya2018unsupervised} & 0.495  & 110.803 & Unsupervised\\
\textsuperscript{\ref{note1}} Hsu et al \cite{hsu2021capturing} & 0.684 & 97.641  & Unsupervised\\ 
\textbf{Ours} &\textbf{0.719} & \textbf{27.838} & Unsupervised\\
\midrule
$\lambda_{inv}=0$ & 0.696 & 38.645 & Unsupervised\\
$\lambda_f=0$ & 0.677 & 42.318 & Unsupervised\\
$\lambda_v=0$ & 0.671 & 41.801 & Unsupervised\\
w/ Med3D \cite{chen2019med3d} features & 0.657 & 29.906 & Unsupervised\\
$k$-means & 0.439 & 63.811 & Non-DL\\
Features $k$-means  & 0.471 & 45.917 & Unsupervised\\
Med3D \cite{chen2019med3d} $k-$means & 0.231 & 55.846 & Unsupervised\\
\bottomrule
\end{tabular}
\caption{\textbf{BraTS'19 results with ablation studies.} Our method outperforms previous unsupervised methods in both the Dice score and the $95\%$ Hausdorff distance.}
\label{bratstable}
\vspace{-2.1em}
\end{table}

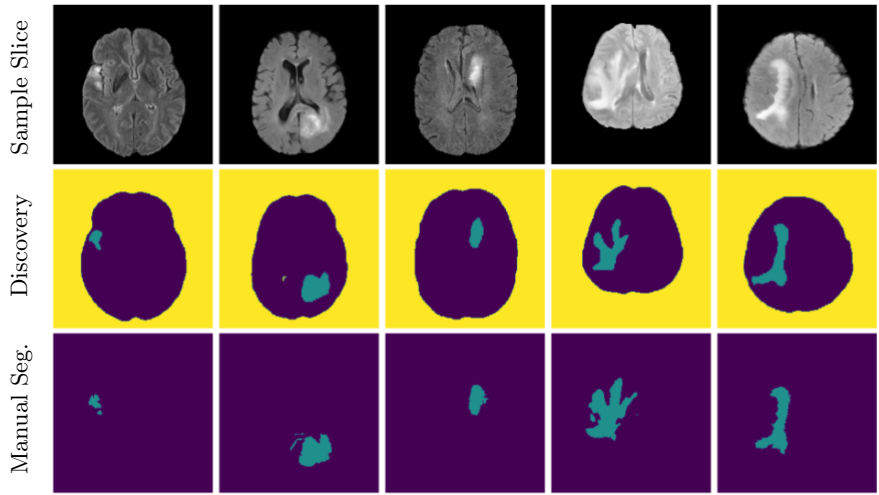
\begin{figure}[t]
\begin{minipage}{0.9\linewidth}
\begin{picture}(200,200)
\put(25,0){\includegraphics[trim=0cm 0cm 2cm 0cm, clip=true,width=\linewidth,right]{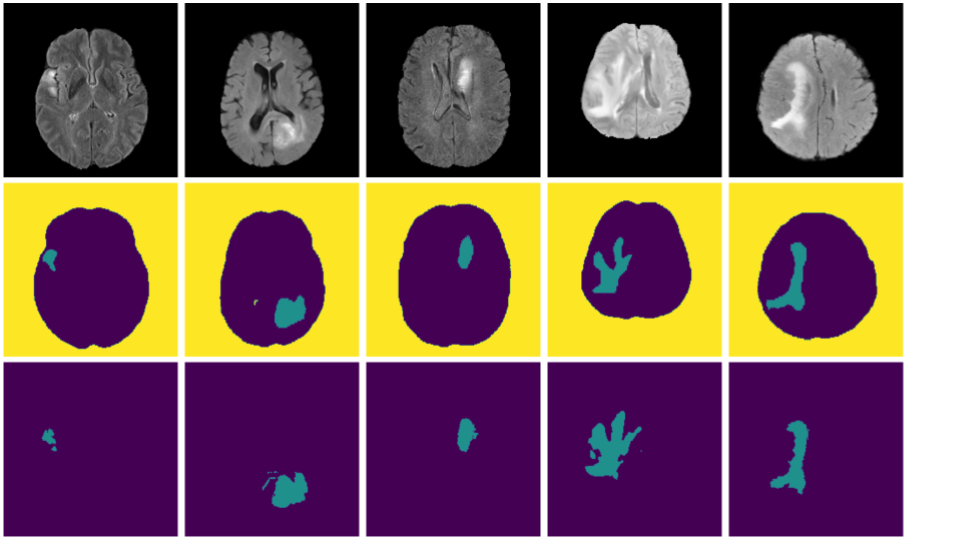}}
\put(10,129){\rotatebox{90}{Sample Slice}}
\put(10,75){\rotatebox{90}{Discovery}}
\put(10,8){\rotatebox{90}{Manual Seg.}}
\end{picture}
\end{minipage}
\caption{\textbf{Examples of discovered structures on BraTS'19.} Our method discovers meaningful regions and detects tumors of different sizes in an  unsupervised manner.}
\vspace{-1.7em}
\label{brats}

\end{figure}

For the Brain Tumor Segmentation (BraTS'19) dataset, we use the whole tumor (WT) segmentation mask for evaluation, which is detectable based on the FLAIR images alone. We train segmentation models with $K = 3$ parts (background, brain, tumor). The evaluation metric, as in the BraTS'19 challenge \cite{menze2014multimodal}, is Dice score and the $95$th percentile of the symmetric Hausdorff distance, which quantifies the surface distance of the predicted segmentation from the manual tumor segmentation in millimeters. Tab.~\ref{bratstable} shows that our model outperforms all prior unsupervised methods for both evaluation metrics. As an approximate upper bound we show for reference the reported results of the $1st$ place solution~\cite{jiang2019two} on BraTS'19 which is based on supervised  training on the full train set and evaluated on the BraTS'19 test set. The qualitative results in Fig.~\ref{brats} show that our model can detect tumors of different sizes. Our predictions look smoother and do not capture fine details of tumor segmentations. 

We perform ablation studies on the BraTS'19 dataset (Tab.~\ref{bratstable}: below the line). Measuring the impact of each loss, we see that the smallest performance drop is due to a deactivated invariance loss ($\lambda_{inv}=0$) while deactivating the visual consistency ($\lambda_v=0$) and feature consistency ($\lambda_f=0$) losses results in larger, but similar performance drops. However, to achieve best performance all three components are necessary. We also perform $k$-means clustering on intensities and features. We observe that useing our deep network model dramatically improves performance, although our losses are similar to $k-$means clustering. This might be due to the fact that predictive modeling involves learning from a distribution of images and a model may therefore extract useful knowledge from a collection of images. To evaluate the significance of the diffusion features, we replaced our diffusion feature extractor with a 3D ResNet from Med3D~\cite{chen2019med3d} trained on 23 medical datasets. We use the ''layer1\_2\_conv2'' features as they showed the best performance.  Although performance does not drop significantly when Med3D features are used with our losses, Med3D features do not produce good results when directly used for $k$-means clustering.\vspace{-1em}

\section{Conclusion}
In this work, we showed that features from 3D generative diffusion models using a ladder-like U-Net-based architecture can discover \emph{intrinsic} 3D structures in biomedical images. We trained predictive \emph{unsupervised} segmentation models using losses that encourage the decomposition of biomedical volumes into nested subvolumes aligned with their hierarchical structures. Our method outperforms existing unuspervised segmentation approaches and discovers meaningful hierarchical concepts on challenging biologically-inspired synthetic datasets and on the BraTS brain tumor dataset. While we tested our approach for unsupervised image segmentation it is conceivable that it could also be useful in semi-supervised settings and that could be applied to data types other than images. 

\textbf{Ackknoledgements:} This work was supported by NIH grants 1R01AR072013, 1R01HL149877, and R41MH118845. The work expresses the views of the authors, not of NIH.

\bibliographystyle{splncs04}
\bibliography{main}

\begin{thebibliography}{10}
\providecommand{\url}[1]{\texttt{#1}}
\providecommand{\urlprefix}{URL }
\providecommand{\doi}[1]{https://doi.org/#1}

\bibitem{asiedu2022decoder}
Asiedu, E.B., Kornblith, S., Chen, T., Parmar, N., Minderer, M., Norouzi, M.:
  Decoder denoising pretraining for semantic segmentation. arXiv:2205.11423
  (2022)

\bibitem{bakas2017advancing}
Bakas, S., Akbari, H., Sotiras, A., Bilello, M., Rozycki, M., Kirby, J.S.,
  Freymann, J.B., Farahani, K., Davatzikos, C.: Advancing the cancer genome
  atlas glioma {MRI} collections with expert segmentation labels and radiomic
  features. Scientific data  \textbf{4}(1),  1--13 (2017)

\bibitem{bakas2018identifying}
Bakas, S., Reyes, M., Jakab, A., Bauer, S., Rempfler, M., Crimi, A., Shinohara,
  R.T., Berger, C., Ha, S.M., Rozycki, M., et~al.: Identifying the best machine
  learning algorithms for brain tumor segmentation, progression assessment, and
  overall survival prediction in the {B}ra{TS} challenge. arXiv:1811.02629
  (2018)

\bibitem{baranchuk2021label}
Baranchuk, D., Rubachev, I., Voynov, A., Khrulkov, V., Babenko, A.:
  Label-efficient semantic segmentation with diffusion models. ICLR  (2021)

\bibitem{chen2019med3d}
Chen, S., Ma, K., Zheng, Y.: Med3{D}: {T}ransfer learning for 3{D} medical
  image analysis. arXiv:1904.00625  (2019)

\bibitem{choudhury2021unsupervised}
Choudhury, S., Laina, I., Rupprecht, C., Vedaldi, A.: Unsupervised part
  discovery from contrastive reconstruction. NeurIPS  \textbf{34},
  28104--28118 (2021)

\bibitem{cciccek20163d}
{\c{C}}i{\c{c}}ek, {\"O}., Abdulkadir, A., Lienkamp, S.S., Brox, T.,
  Ronneberger, O.: 3{D} {U}-{Net}: learning dense volumetric segmentation from
  sparse annotation. In: MICCAI. pp. 424--432 (2016)

\bibitem{collins2018deep}
Collins, E., Achanta, R., Susstrunk, S.: Deep feature factorization for concept
  discovery. In: ECCV. pp. 336--352 (2018)

\bibitem{donahue2019large}
Donahue, J., Simonyan, K.: Large scale adversarial representation learning.
  NeurIPS  \textbf{32} (2019)

\bibitem{dorjsembe2022three}
Dorjsembe, Z., Odonchimed, S., Xiao, F.: Three-dimensional medical image
  synthesis with denoising diffusion probabilistic models. In: MIDL (2022)

\bibitem{gao2017classification}
Gao, X.W., Hui, R., Tian, Z.: Classification of {CT} brain images based on deep
  learning networks. Computer methods and programs in biomedicine
  \textbf{138},  49--56 (2017)

\bibitem{ho2020denoising}
Ho, J., Jain, A., Abbeel, P.: Denoising diffusion probabilistic models. NeurIPS
   \textbf{33},  6840--6851 (2020)

\bibitem{hsu2021capturing}
Hsu, J., Gu, J., Wu, G., Chiu, W., Yeung, S.: Capturing implicit hierarchical
  structure in 3{D} biomedical images with self-supervised hyperbolic
  representations. NeurIPS  \textbf{34},  5112--5123 (2021)

\bibitem{hu2022unsupervised}
Hu, D., Tao, Y.K., Oguz, I.: Unsupervised denoising of retinal {OCT} with
  diffusion probabilistic model. In: Medical Imaging 2022: Image Processing.
  vol. 12032, pp. 25--34 (2022)

\bibitem{hung2019scops}
Hung, W.C., Jampani, V., Liu, S., Molchanov, P., Yang, M.H., Kautz, J.:
  S{COPS}: Self-supervised co-part segmentation. In: CVPR. pp. 869--878 (2019)

\bibitem{ji2019invariant}
Ji, X., Henriques, J.F., Vedaldi, A.: Invariant information clustering for
  unsupervised image classification and segmentation. In: ICCV. pp. 9865--9874
  (2019)

\bibitem{jiang2019two}
Jiang, Z., Ding, C., Liu, M., Tao, D.: Two-stage cascaded {U}-{N}et: 1st place
  solution to {B}ra{TS} challenge 2019 segmentation task. In: International
  MICCAI brain lesion workshop. pp. 231--241 (2019)

\bibitem{kim2022diffusion}
Kim, B., Oh, Y., Ye, J.C.: Diffusion adversarial representation learning for
  self-supervised vessel segmentation. arXiv:2209.14566  (2022)

\bibitem{li2021semantic}
Li, D., Yang, J., Kreis, K., Torralba, A., Fidler, S.: Semantic segmentation
  with generative models: Semi-supervised learning and strong out-of-domain
  generalization. In: CVPR. pp. 8300--8311 (2021)

\bibitem{maggioni2012nonlocal}
Maggioni, M., Katkovnik, V., Egiazarian, K., Foi, A.: Nonlocal transform-domain
  filter for volumetric data denoising and reconstruction. IEEE TIP
  \textbf{22}(1),  119--133 (2012)

\bibitem{menze2014multimodal}
Menze, B.H., Jakab, A., Bauer, S., Kalpathy-Cramer, J., Farahani, K., Kirby,
  J., Burren, Y., Porz, N., Slotboom, J., Wiest, R., et~al.: The multimodal
  brain tumor image segmentation benchmark ({B}ra{TS}). IEEE TMI
  \textbf{34}(10),  1993--2024 (2014)

\bibitem{milletari2016v}
Milletari, F., Navab, N., Ahmadi, S.A.: V-{N}et: Fully convolutional neural
  networks for volumetric medical image segmentation. In: International
  Conference on 3D Vision (3DV). pp. 565--571 (2016)

\bibitem{moriya2018unsupervised}
Moriya, T., Roth, H.R., Nakamura, S., Oda, H., Nagara, K., Oda, M., Mori, K.:
  Unsupervised segmentation of 3{D} medical images based on clustering and deep
  representation learning. In: Medical Imaging 2018: Biomedical Applications in
  Molecular, Structural, and Functional Imaging. vol. 10578, pp. 483--489
  (2018)

\bibitem{nalepa2020unsupervised}
Nalepa, J., Myller, M., Imai, Y., Honda, K., Takeda, T., Antoniak, M.:
  Unsupervised segmentation of hyperspectral images using 3-{D} convolutional
  autoencoders. IEEE Geoscience and Remote Sensing Letters  \textbf{17}(11),
  1948--1952 (2020)

\bibitem{nalepa2019data}
Nalepa, J., Marcinkiewicz, M., Kawulok, M.: Data augmentation for brain-tumor
  segmentation: a review. Frontiers in Computational Neuroscience  \textbf{13},
  ~83 (2019)

\bibitem{park2022encoder}
Park, J., Yang, H., Roh, H.J., Jung, W., Jang, G.J.: Encoder-weighted {W-Net}
  for unsupervised segmentation of cervix region in colposcopy images. Cancers
  \textbf{14}(14), ~3400 (2022)

\bibitem{pinaya2022fast}
Pinaya, W.H., Graham, M.S., Gray, R., Da~Costa, P.F., Tudosiu, P.D., Wright,
  P., Mah, Y.H., MacKinnon, A.D., Teo, J.T., Jager, R., et~al.: Fast
  unsupervised brain anomaly detection and segmentation with diffusion models.
  In: MICCAI. pp. 705--714 (2022)

\bibitem{ronneberger2015u}
Ronneberger, O., Fischer, P., Brox, T.: U-{N}et: Convolutional networks for
  biomedical image segmentation. In: MICCAI. pp. 234--241. Springer (2015)

\bibitem{sanchez2022healthy}
Sanchez, P., Kascenas, A., Liu, X., O’Neil, A.Q., Tsaftaris, S.A.: What is
  healthy? generative counterfactual diffusion for lesion localization. In:
  DGM4MICCAI. pp. 34--44 (2022)

\bibitem{sejdic2013necessity}
Sejdi{\'c}, E., Lipsitz, L.A.: Necessity of noise in physiology and medicine.
  Computer methods and programs in biomedicine  \textbf{111}(2),  459--470
  (2013)

\bibitem{wolleb2022diffusion}
Wolleb, J., Bieder, F., Sandk{\"u}hler, R., Cattin, P.C.: Diffusion models for
  medical anomaly detection. In: MICCAI. pp. 35--45 (2022)

\bibitem{wolleb2022diffusion1}
Wolleb, J., Sandk{\"u}hler, R., Bieder, F., Valmaggia, P., Cattin, P.C.:
  Diffusion models for implicit image segmentation ensembles. In: MIDL. pp.
  1336--1348 (2022)

\bibitem{wu2022medsegdiff}
Wu, J., Fang, H., Zhang, Y., Yang, Y., Xu, Y.: {MedSegDiff}: Medical image
  segmentation with diffusion probabilistic model. arXiv:2211.00611  (2022)

\bibitem{wyatt2022anoddpm}
Wyatt, J., Leach, A., Schmon, S.M., Willcocks, C.G.: Anoddpm: Anomaly detection
  with denoising diffusion probabilistic models using simplex noise. In: CVPR.
  pp. 650--656 (2022)

\bibitem{yang2017quicksilver}
Yang, X., Kwitt, R., Styner, M., Niethammer, M.: Quicksilver: Fast predictive
  image registration--a deep learning approach. NeuroImage  \textbf{158},
  378--396 (2017)

\bibitem{zhao2019data}
Zhao, A., Balakrishnan, G., Durand, F., Guttag, J.V., Dalca, A.V.: Data
  augmentation using learned transformations for one-shot medical image
  segmentation. In: CVPR. pp. 8543--8553 (2019)

\end{thebibliography}
\end{document}